\documentclass[conference]{IEEEtran}

\usepackage{graphicx}
\usepackage{amsmath}
\usepackage{booktabs}
\usepackage{array}
\usepackage{url}
\usepackage{cite}
\usepackage{balance}
\usepackage{multirow}

\title{Bee Detection and Tracking at Hive Entrance using YOLO11 and ByteTrack}

\author{
\IEEEauthorblockN{Thi Thu Thao Nguyen}
\IEEEauthorblockA{
IoT Engineering\\
Savonia University of Applied Sciences\\
Finland\\
Thaonguyenthithu0311@gmail.com}
\and
\IEEEauthorblockN{Johannes Reschke}
\IEEEauthorblockA{
Electrical and Information Engineering\\
Ostbayerische Technische Hochschule Regensburg\\
Germany\\
johannes.reschke@oth-regensburg.de}
}

\begin{document}
\maketitle

\begin{abstract}
This work presents an automatic bee entrance monitoring system based on YOLO11 transfer learning and the ByteTrack tracking algorithm. The study investigates the influence of data augmentation, backbone freezing, and tracker parameter optimization on the detection and counting of small, fast-moving bees. The detector with progressive backbone unfreezing strategy achieved about 97.0\% precision and 98.7\% mAP50, while providing more stable convergence than full fine-tuning. Experiments also showed that light augmentation outperformed heavy augmentation. For tracking, ByteTrack parameters were optimized to improve trajectory continuity under low-confidence detections. On an independent 25 FPS side-view video, the optimized YOLO11-ByteTrack system correctly counted 43 of 47 incoming bees (91.5\%) and 7 of 30 outgoing bees (23.3\%). Error analysis showed that most counting errors were caused by missed detections due to rapid bee motion and motion blur, while tracking failures became less frequent after parameter optimization. Overall, the results indicate that moderate augmentation, progressive backbone unfreezing, and ByteTrack tuning improve the reliability of automatic bee entrance monitoring under realistic recording conditions.
\end{abstract}


\section{Introduction}
Monitoring bee traffic at hive entrances has become an important task in precision apiculture. The number of bees entering and leaving a hive reflects colony activity and may indicate diseases, food availability, swarming behaviour, or pesticide exposure.

Traditional manual counting is labor-intensive and unsuitable for continuous monitoring. Recent advances in deep learning enable automatic bee detection and tracking using computer vision.

However, tracking bees near the hive entrance is still difficult. Bees occupy only a small part of the image, move rapidly, and are frequently affected by motion blur. In addition, several bees may overlap near the entrance, which can lead to fragmented trajectories and ID switches. This work investigates how modern object detection and tracking algorithms perform under these challenging conditions.

This work focuses on the following aspects:
\begin{itemize}
    \item training a YOLO11 detector for bee detection,
    \item comparing different augmentation strategies,
    \item evaluating backbone freezing schemes,
    \item optimizing ByteTrack parameters for bee tracking,
    \item analyzing the main sources of counting errors.
\end{itemize}

\section{Related Work}
Recent studies have shown that combining deep learning object detection with multi-object tracking is a practical solution for automatic bee monitoring at hive entrances. A closely related study, \emph{A Honey Bee In-and-Out Counting Method Based on Multiple Object Tracking Algorithm}~\cite{bee_counting}, proposed a complete pipeline using YOLOv8m together with ByteTrack, OC-SORT, and Deep OC-SORT. The authors also compared a virtual counting line with a counting box and reported the best performance with OC-SORT and the box-based strategy, achieving F1-scores of 91.49\% for incoming bees and 89.08\% for outgoing bees. The study provides a strong baseline for bee entrance counting. However, the experiments were conducted on videos collected from only two hives, and the training and testing data came from the same environment. This makes it difficult to judge how well the system would perform under different viewpoints, backgrounds, or lighting conditions. In addition, the impact of tracker parameter optimization was not investigated.

Another relevant work is \emph{A Method for Bee Activities Recognition from Videos Captured at the Beehive Entrance}~\cite{bee_activity}, which combined YOLOv5 with OC-SORT to recognize behaviors such as entering, leaving, and carrying pollen. The authors reported high detection precision and good tracking performance. The main limitation of this work is that it focuses on behavior recognition rather than counting accuracy. In addition, only one detector and one tracker were evaluated, and the effect of tracker parameter settings was also not investigated.

For long-term monitoring, \emph{Continuous Non-Invasive Monitoring of Hive Entrance Activity Reveals Honey Bee Colony Dynamics}~\cite{hive_dynamics} used YOLOv8 together with BoT-SORT and evaluated the system on videos that were independent from the training data. This is an important strength because it provides a more realistic assessment of generalization. Nevertheless, the study concentrates on continuous colony monitoring and does not provide a detailed analysis of counting errors such as missed detections, ID switches, or temporary tracking loss near the hive entrance.

Recent work has also emphasized the importance of video quality. \emph{Physics-aware Vision Instrumentation for Stingless Bee Counting at Hive Entrance Using Hybrid Edge-Cloud Object Detection}~\cite{physics_aware} showed that lower frame rates increase the displacement of bees between consecutive frames, which makes tracking more difficult and increases the probability of missed detections and ID switches. While the study offers valuable insight into the relationship between hardware constraints and tracking performance, it focuses on stingless bees and does not evaluate newer detectors such as YOLO11.

Motivated by these limitations, the present study uses publicly available bee detection datasets together with an independent tracking video recorded from a different viewpoint. It evaluates YOLO11 detector combined with ByteTrack, investigates the influence of tracker parameter optimization, and provides a detailed analysis of detection failures, tracking failures, and counting errors under more challenging and realistic conditions.

\section{Dataset}

\subsection{Detection Dataset}
The dataset combines two publicly available sources: the Mendeley dataset~\cite{mendeley} containing over 6,500 images and Dataset Ninja~\cite{dataset_ninja} containing over 3,000 images. In total, there are over 9,500 images, split into 7,635 training images and 1,909 validation images.

The Mendeley dataset provides a large collection of labeled images focusing on bee detection and movement direction on beehive landing boards, while the Dataset Ninja collection contributes additional object detection images with diverse visual characteristics. They together enhance the quantity and variability in appearance, viewpoint, and environmental conditions.

\begin{figure}[t]
\centering
\includegraphics[width=\columnwidth]{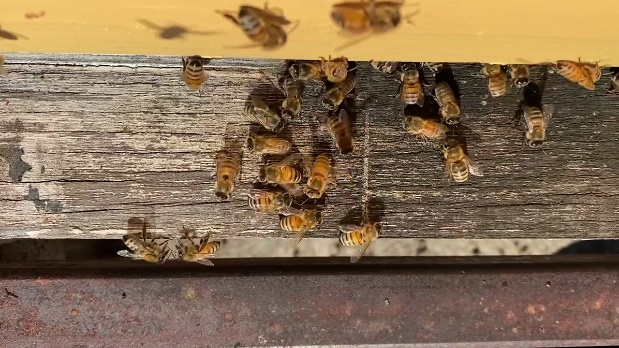}
\caption{Example image of bees at a hive entrance, adapted from the Mendeley dataset~\cite{mendeley}.}
\label{fig:mendeley1}
\end{figure}

\begin{figure}[t]
\centering
\includegraphics[width=\columnwidth]{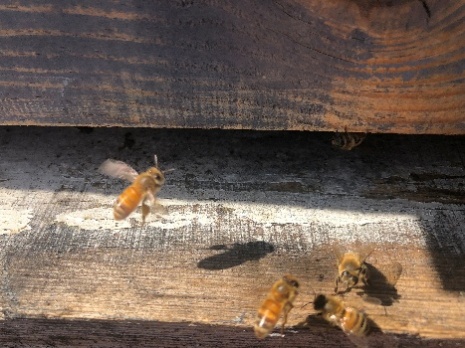}
\caption{Example image of bees at a hive entrance, adapted from the Mendeley dataset~\cite{mendeley}.}
\label{fig:mendeley2}
\end{figure}

\begin{figure}[t]
\centering
\includegraphics[width=\columnwidth]{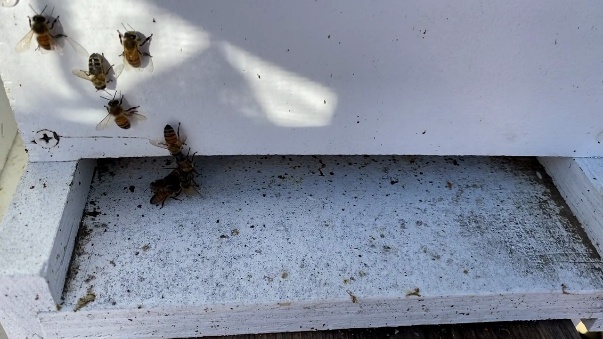}
\caption{Example image of bees at a hive entrance, adapted from Dataset Ninja~\cite{dataset_ninja}.}
\label{fig:datasetninja}
\end{figure}

\subsection{Tracking Evaluation Video}
The evaluation video was collected from Pexels~\cite{pexels}. It was recorded at 25 FPS and 30 seconds long, using a side-view camera under natural lighting conditions. Because this video was not included in the training data, it was used to evaluate the model's ability to generalize to a different scene rather than to memorize the training environment.

\begin{figure}[t]
\centering
\includegraphics[width=\columnwidth]{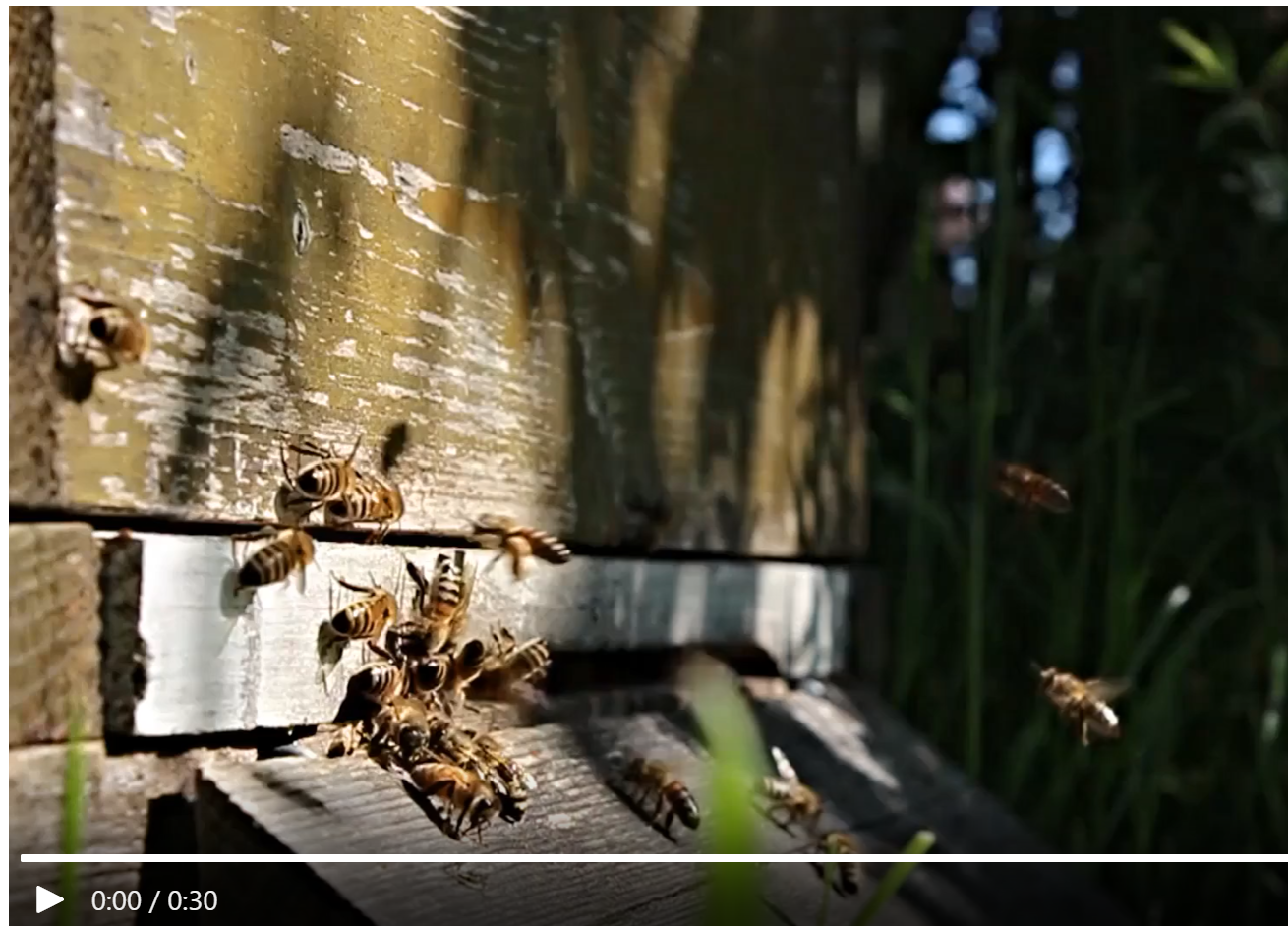}
\caption{A frame extracted from the evaluation video by Ella Gronewold on Pexels~\cite{pexels}.}
\label{fig:video}
\end{figure}

\section{Methodology}

\subsection{Detection Model and Training Strategy}

\subsubsection{YOLOv8 and YOLO11}
This study evaluates YOLOv8 and YOLO11 for bee detection. Both are one-stage, anchor-free object detectors that provide high detection accuracy with real-time inference speed. Compared with YOLOv8, YOLO11 offers improved feature representation and localization performance, making it more suitable for detecting small, fast-moving bees~\cite{yolo_comparison}.

\subsubsection{Transfer Learning}
Both models were initialized with COCO pre-trained weights and fine-tuned on the bee dataset. Transfer learning accelerates convergence, improves detection accuracy with limited data, and reduces the risk of overfitting.

\subsection{Tracking}
After object detection, detected bees are tracked across consecutive frames using the ByteTrack multi-object tracking algorithm. Beside associating high-confidence detections, ByteTrack also considers low-confidence detection boxes. By matching both high- and low-confidence detections with existing trajectories, the algorithm recovers temporarily weak detections and produces more continuous object tracks~\cite{bytetrack}.

\begin{figure}[t]
\centering
\includegraphics[width=.78\columnwidth]{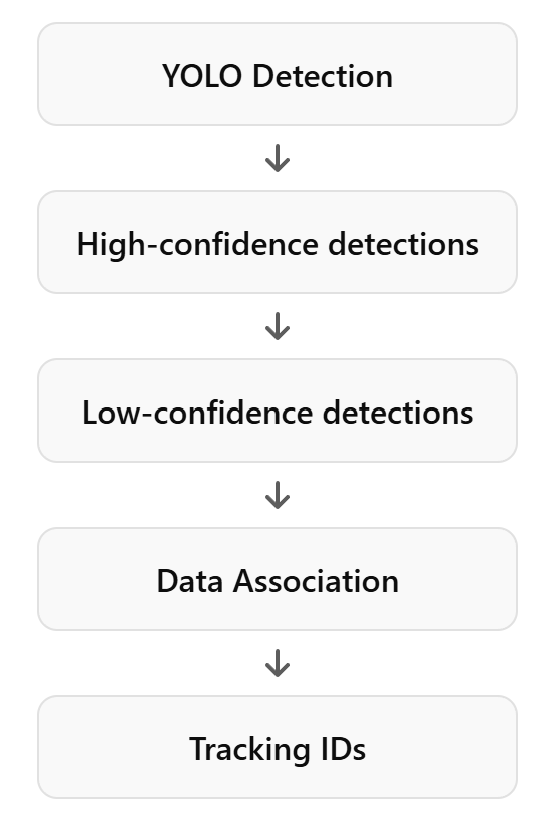}
\caption{ByteTrack pipeline.}
\label{fig:bytetrack}
\end{figure}

The tracking pipeline is illustrated in Fig.~\ref{fig:bytetrack}.

ByteTrack was selected because honey bees are small, fast-moving objects that often produce low-confidence detections due to motion blur and viewpoint changes. By retaining and associating these low-confidence detections instead of discarding them, ByteTrack reduces missed trajectories and improves counting accuracy while maintaining real-time performance.

\subsection{Counting Logic}
The final stage of the proposed system estimates the number of bees entering and leaving the hive based on their tracked trajectories.

\begin{figure}[t]
\centering
\includegraphics[width=.78\columnwidth]{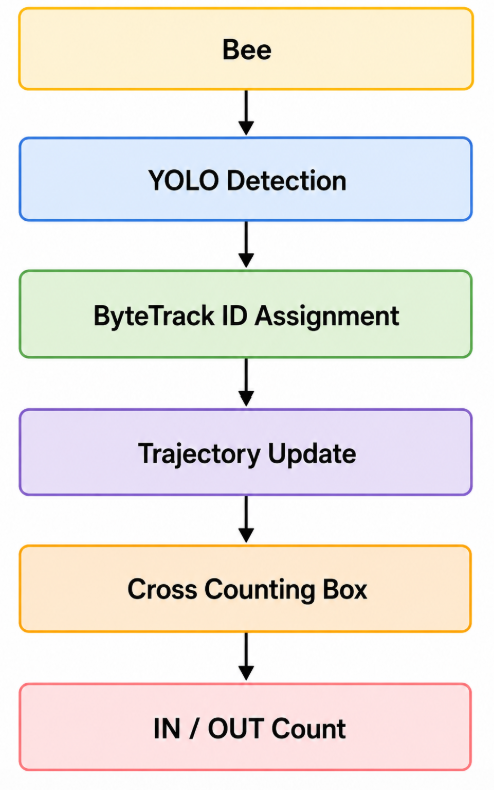}
\caption{Counting logic graph.}
\label{fig:countinglogic}
\end{figure}

A rectangular counting box is defined at the hive entrance. For each tracked bee, the center point of its bounding box is computed in every frame. The current center position is then compared with its position in the previous frame to determine whether the bee has crossed the boundary of the counting box.

A bee is counted as IN or OUT when its center point crosses the bounding box boundary from the outside in, or from the inside out, respectively. Each counted tracking ID is added to the IN/OUT log for reference. If a bee enters and exits multiple times, each occurrence is recorded independently.

Compared with a single virtual counting line, the counting-box approach is more robust to the oscillatory flight behavior of bees near the hive entrance.

\section{Experiments}

\subsection{Experiment 1: Effect of Augmentation}
Data augmentation is important to model performance under limited-data conditions. However, when it comes to small objects like bees, our observation shows that heavy augmentation with RandAugment and random erasing can substantially diminish model performance in terms of mAP50--95, box loss, class loss, and recall. Meanwhile, light augmentation consistently achieves higher localization accuracy, higher recall, and lower losses, suggesting an optimization in model learning~\cite{small_object_aug}.

\begin{figure}[t]
\centering
\includegraphics[width=\columnwidth]{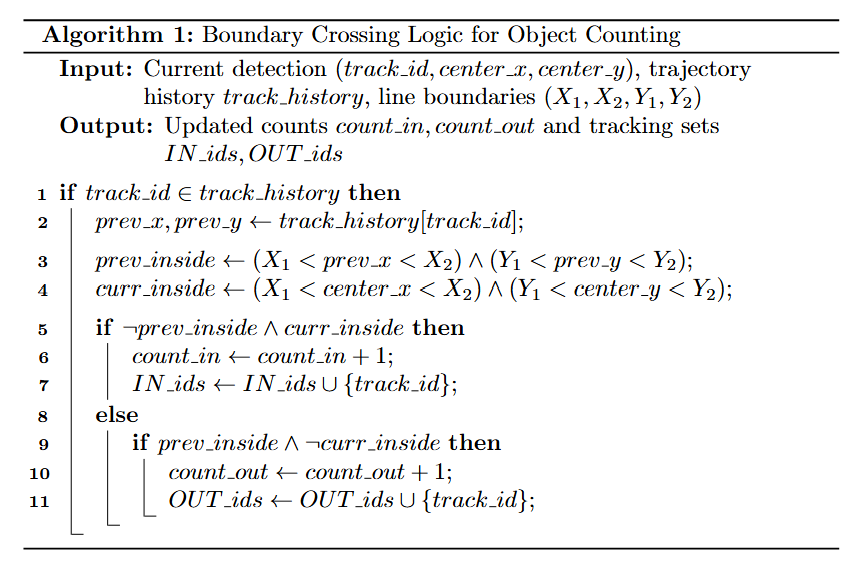}
\caption{Box loss and class loss between light and heavy augmented models.}
\label{fig:loss_aug}
\end{figure}

\begin{figure}[t]
\centering
\includegraphics[width=\columnwidth]{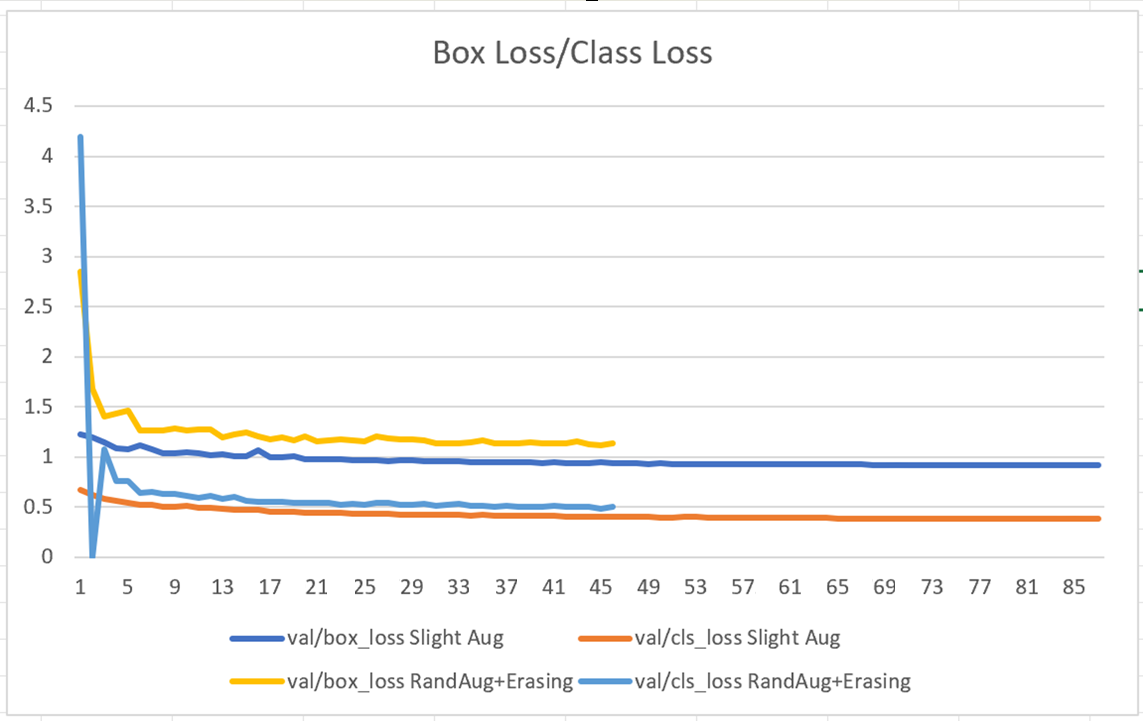}
\caption{mAP50 and mAP50--95 between light and heavy augmented models.}
\label{fig:map_aug}
\end{figure}

\begin{figure}[t]
\centering
\includegraphics[width=\columnwidth]{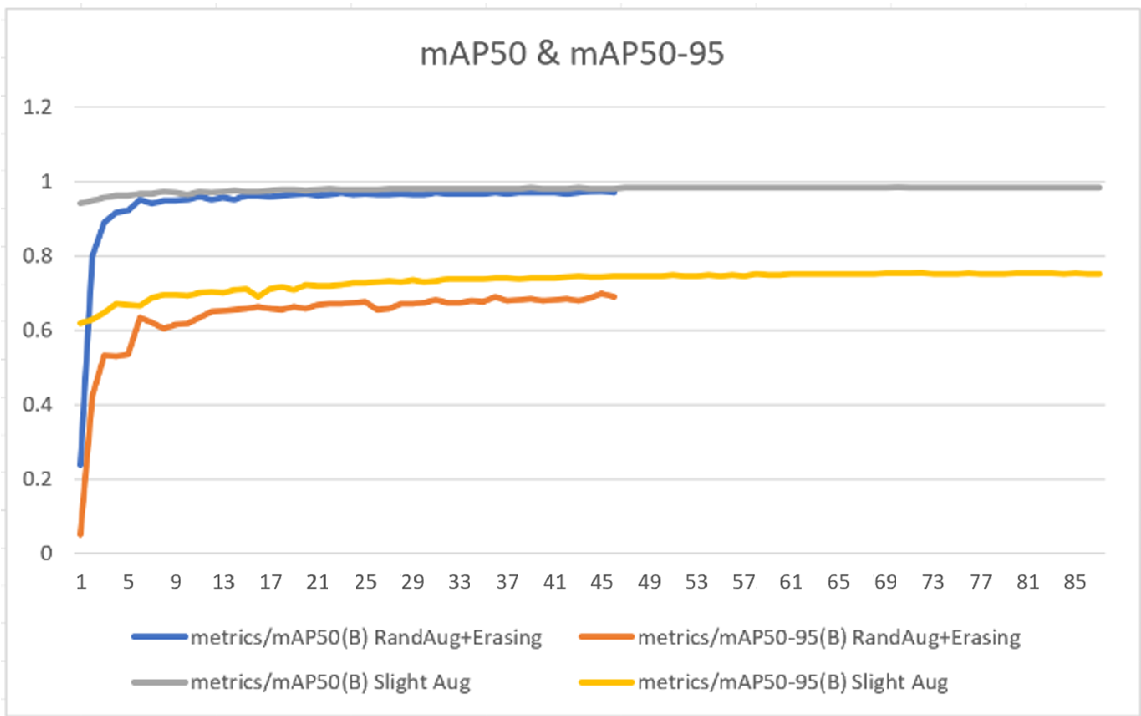}
\caption{Recall between light and heavy augmented models.}
\label{fig:recall_aug}
\end{figure}

Slight augmentation includes model \texttt{yolov8s.pt}, \texttt{imgsz=640}, \texttt{close\_mosaic=15}, \texttt{hsv\_h=0.01}, \texttt{hsv\_s=0.25}, \texttt{hsv\_v=0.25}, \texttt{degrees=3.0}, \texttt{translate=0.04}, \texttt{scale=0.08}, \texttt{fliplr=0.3}, \texttt{mosaic=0.3}, and \texttt{mixup=0.05}. Heavy augmentation includes slight augmentation plus RandAugment.

The reason lies in the fact that bees typically occupy less than 1--3\% of the image area. Once random augmentation is applied, some shearing or erasing can significantly transform the bees, make them distorted, and cause them to lose critical visual details. This becomes noise in the features and negatively impacts detection accuracy.

\subsection{Experiment 2: Backbone Freezing}
The following tables summarize the performance of three training strategies: fully unfreezing all layers (fine-tune all), freezing the backbone for the first 30 epochs and then unfreezing (progressive backbone unfreezing), and freezing the backbone throughout the entire training process~\cite{finetune_guide}. While all configurations achieve comparably high precision and recall values, there are differences in validation losses, convergence behavior, and training stability.

\begin{table}[t]
\centering
\caption{Precision and Recall Comparison}
\label{tab:precision_recall}
\small
\begin{tabular}{lcc}
\toprule
Model & Precision & Recall\\
\midrule
Unfreeze all & $\sim$0.969--0.973 & $\sim$0.960--0.964\\
Freeze backbone 30 epochs & $\sim$0.971--0.974 & $\sim$0.958--0.963\\
Freeze backbone all epochs & $\sim$0.964--0.968 & $\sim$0.958--0.961\\
\bottomrule
\end{tabular}
\end{table}

\begin{table}[t]
\centering
\caption{mAP50 and mAP50--95 Comparison}
\label{tab:map}
\small
\begin{tabular}{lcc}
\toprule
Model & mAP50 & mAP50--95\\
\midrule
Unfreeze all & $\sim$0.984 & $\sim$0.752--0.754\\
Freeze backbone 30 epochs & $\sim$0.987 & $\sim$0.783--0.784\\
Freeze backbone all epochs & $\sim$0.986 & $\sim$0.783--0.784\\
\bottomrule
\end{tabular}
\end{table}

\begin{table}[t]
\centering
\caption{Validation Losses Comparison}
\label{tab:losses}
\small
\begin{tabular}{lccc}
\toprule
Model & box loss & cls loss & dfl loss\\
\midrule
Unfreeze all & $\sim$0.875 & $\sim$0.345 & $\sim$1.018\\
Freeze backbone 30 epochs & $\sim$0.862 & $\sim$0.360 & $\sim$0.955\\
Freeze backbone all epochs & $\sim$0.933 & $\sim$0.380 & $\sim$0.960\\
\bottomrule
\end{tabular}
\end{table}

\begin{table}[t]
\centering
\caption{Training Duration}
\label{tab:duration}
\small
\begin{tabular}{lc}
\toprule
Model & Epochs\\
\midrule
Unfreeze all & 81\\
Freeze backbone 30 epochs & 93 (highest)\\
Freeze backbone all epochs & 75\\
\bottomrule
\end{tabular}
\end{table}

When comparing the three backbone training strategies, mAP50, mAP50--95 and losses over epochs reveal distinct performance. With mAP50--95 around 0.75--0.753, freezing the backbone for all epochs results in the lowest overall performance, showing minimal localization accuracy. Box loss, classification loss and DFL loss remain highest among the three strategies, implying limited adaptability to the target dataset. This is due to features in the backbone being fixed throughout training.

Full fine-tuning achieves high mAP50--95 at 0.784--0.785 along with over 96\% precision and recall, demonstrating strong localization capability. However, its performance fluctuates more across epochs and is accompanied by higher validation losses as compared to progressive backbone unfreezing by over 6\%, suggesting a greater risk of overfitting.

In contrast, progressive backbone freezing delivers the most stable and balanced performance. With high mAP50 ($\approx$0.986--0.987), highest precision ($\approx$0.971--0.974) and around 96\% recall rate, it represents accuracy and robustness. Additionally, this strategy gains the lowest validation DFL loss ($\approx$0.955), indicating more accurate bounding box distribution and improved localization accuracy.

From a convergence aspect, the unfreezing-all model reaches early convergence at epoch 82, whereas the progressive backbone unfreezing, which freezes the backbone for the first 30 epochs, trains for the longest duration (93 epochs), indicating improved training stability and delayed overfitting. Given a small and less diverse dataset, full fine-tuning increases the risk of rapid memorization of training samples rather than learning generalizable representations.

Overall, under limited-data conditions, progressive unfreezing, specifically freezing the backbone during early epochs, provides the best stability and task-specific learning. This strategy effectively mitigates overfitting while improving localization accuracy, making it particularly suitable for small-scale object detection datasets.

\begin{figure}[t]
\centering
\includegraphics[width=\columnwidth]{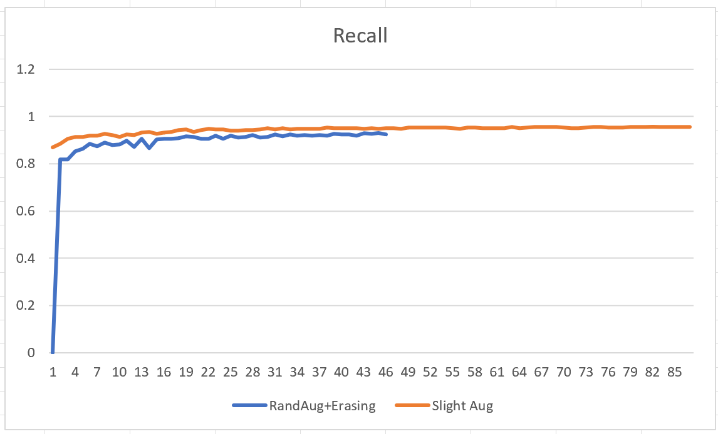}
\caption{Training and validation DFL loss between fine-tune-all and progressive backbone unfreezing.}
\label{fig:dfl}
\end{figure}

\subsection{Experiment 3: Tracking Performance using YOLOv8 vs YOLO11}
This experiment compares the tracking performance of YOLOv8 and YOLO11 using the same tracking pipeline.

YOLO11 preserved bee features more effectively, especially when bees moved quickly or were partially occluded. The C2PSA module improved attention to small localized targets, resulting in fewer missed detections~\cite{yolo_architecture}.

Tracking was also more stable with YOLO11. The use of C3k2 blocks produced more consistent features across consecutive frames, which reduced ID switches during tracking. As a result, YOLO11 achieved better tracking continuity and higher counting accuracy in continuous hive monitoring.

\subsection{Experiment 4: Tracker Parameter Optimization}
The optimized ByteTrack parameters and their rationale are summarized in Table~\ref{tab:bytetrack_params}.

\begin{table}[t]
\centering
\caption{Optimized ByteTrack Tracking Parameters and Their Rationale}
\label{tab:bytetrack_params}
\scriptsize
\begin{tabular}{p{0.23\columnwidth}p{0.18\columnwidth}p{0.49\columnwidth}}
\toprule
Parameter & Selected value & Rationale\\
\midrule
\texttt{track\_high\_thresh} & 0.15, lower than default 0.25 &
Detection confidence fluctuates because of small size and fast motion. Lowering the threshold allows more valid detections to participate in the first association stage.\\
\texttt{track\_low\_thresh} & 0.03, lower than default 0.10 &
A very low threshold enables recovery of weak detections during the second association stage instead of terminating the track.\\
\texttt{new\_track\_thresh} & 0.15, lower than default 0.25 &
Allows newly appearing bees to be initialized more quickly, reducing missed bee entries.\\
\texttt{track\_buffer} & 50, higher than default 30 &
Keeps lost tracks alive longer, enabling successful re-association after temporary occlusion and reducing fragmented trajectories.\\
\texttt{match\_thresh} & 0.85, higher than default 0.80 &
Requires stronger spatial consistency before assigning an existing ID, reducing incorrect ID switches between nearby bees.\\
\texttt{fuse\_score} & True (default) &
Combines detection confidence with motion similarity during association, improving matching robustness.\\
\bottomrule
\end{tabular}
\end{table}

The tracker parameters~\cite{tracking_docs} were intentionally adjusted to prioritize tracking continuity rather than conservative filtering. Honey bees are small, fast-moving, and frequently occluded at the hive entrance, causing temporary drops in detection confidence. Therefore, lower confidence thresholds and a longer track buffer were adopted to preserve object identities across short detection failures.

However, these more permissive settings increase the risk of incorrect associations. To compensate, a higher matching threshold was used, requiring stronger spatial consistency before assigning detections to existing tracks. This combination improves the continuity of bee trajectories while limiting unnecessary ID switches, ultimately leading to more reliable bee counting when individuals cross the virtual counting region.

\section{Results}
The counting performance of YOLOv8 and YOLOv11 under two conditions, with and without tracking parameter optimization, are shown in the graphs as compared to the ground truth.

In the testing video, there are actually 47 bees going in and 30 bees going out of the hive.

\begin{figure}[t]
\centering
\includegraphics[width=\columnwidth]{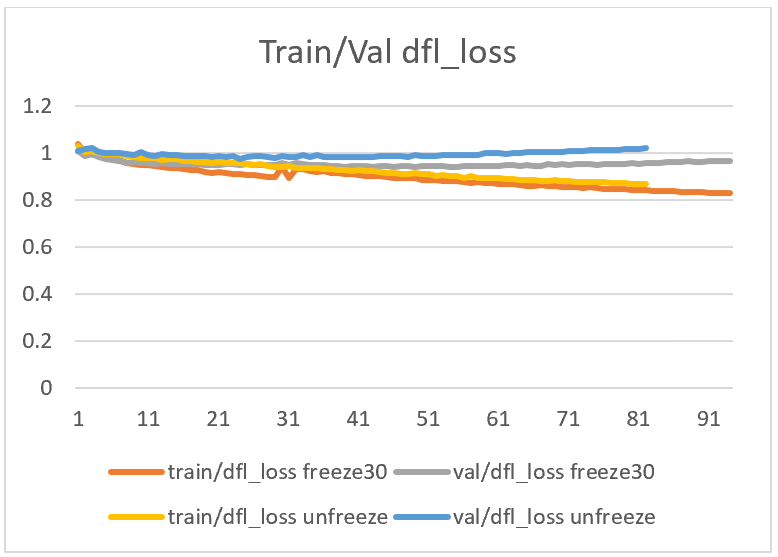}
\caption{Bee counting IN versus ground truth.}
\label{fig:count_in}
\end{figure}

\begin{figure}[t]
\centering
\includegraphics[width=\columnwidth]{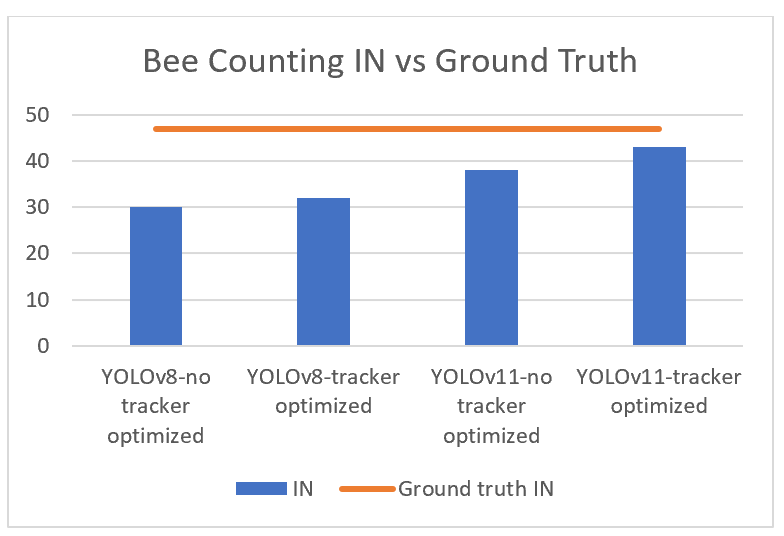}
\caption{Bee counting OUT versus ground truth.}
\label{fig:count_out}
\end{figure}

YOLOv11 consistently outperformed YOLOv8 in both configurations.

Without tracking:
\begin{itemize}
    \item IN: 38 vs. 30 (26.7\% improvement)
    \item OUT: 5 vs. 0
\end{itemize}

With the optimized tracker:
\begin{itemize}
    \item IN: 43 vs. 32 (34.4\% improvement)
    \item OUT: 7 vs. 4 (75\% improvement)
\end{itemize}

The better accuracy of YOLOv11 suggests that it produced more reliable detections for small, fast-moving bees. Better detections provide higher-quality inputs to the tracker, resulting in more stable trajectories and improved counting performance.

Lowering the tracking confidence threshold increased the number of recovered detections, reducing missed bee trajectories. The overall counting accuracy improved. There is also a significant improvement in outgoing bees since low FPS video leads to big displacement when a bee moves fast between frames, hence the model could not regard the bee as the same ID.

\section{Error Analysis}

\subsection{Detection Failures}
The majority of detection failures occurred for outgoing bees, mainly because they move much faster than incoming bees and are captured from more challenging viewing angles. Several factors contributed to these missed detections:

\textbf{Motion blur:} Outgoing bees often leave the hive at high speed, causing motion blur in video frames. This reduces the visibility of important visual features and makes detection more difficult.

\textbf{Viewpoint mismatch:} The training dataset did not contain sufficient variation in bee flight orientations. Bees flying at unusual or side angles may appear significantly different from the top-down views that dominate the dataset, which reduces the model's generalization ability.

\textbf{Lighting variation:} The test video was recorded under relatively stable natural lighting. Illumination had only a minor influence on detection performance, although bees occasionally became darker when flying through shaded areas. Other than that, the recording conditions were relatively stable, so lighting was not considered a major source of error.

\textbf{Bee shadows:} Although shadows were present around the hive entrance, the detection model was generally able to focus on the characteristic body pattern and shape of the bees. As a result, shadows rarely caused false detections.

Overall, the results suggest that high-speed motion and insufficient viewpoint diversity in the training dataset were the primary causes of missed detections, especially for outgoing bees, whereas lighting changes and shadows had relatively little impact on detection performance.

\begin{figure}[t]
\centering
\includegraphics[width=\columnwidth]{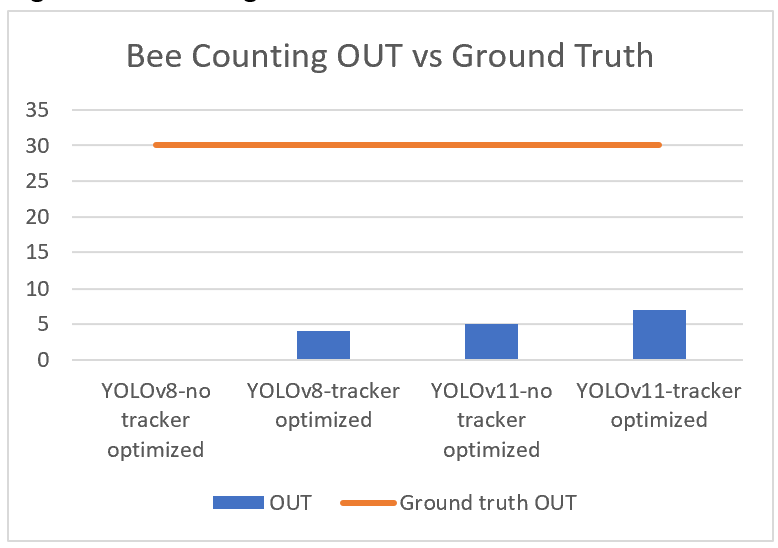}
\caption{Example of a fast-moving bee exiting the hive, demonstrating motion blur and an oblique viewing angle.}
\label{fig:motionblur}
\end{figure}

\subsection{Tracking Failures}
Tracking failures mainly occurred when bees moved quickly between consecutive frames.

\textbf{Fast flight and low FPS:} Because the video was recorded at 25 FPS, fast-moving bees often changed position considerably between frames. Large displacements make it harder for ByteTrack to associate detections with the correct existing track~\cite{bytetrack}.

\textbf{ID switches:} ID switches were observed occasionally, particularly when several bees were close to each other near the entrance.

\textbf{Temporary disappearance:} Bees were sometimes lost for one or two frames because of partial occlusion or reduced detection confidence, but this happened only rarely.

\textbf{Rapid turning:} Sudden changes in flight direction could also break trajectory continuity.

After parameter optimization, the number of fragmented tracks decreased, although a trade-off remained between preserving tracks for fast-moving bees and avoiding incorrect ID associations.

\section{Conclusion}
This study presented a bee entrance monitoring system based on YOLO11 for detection and ByteTrack for tracking. The experiments showed that moderate data augmentation and progressive backbone unfreezing provided the best balance between accuracy and generalization. Tuning the ByteTrack parameters also improved trajectory continuity and reduced counting errors compared with the default configuration.

The error analysis indicated that the main limitation of the current system is missed detection of fast-moving outgoing bees. Once detections are lost, tracking becomes more difficult and counting accuracy decreases. In contrast, incoming bees are generally easier to detect and track because they move more slowly near the hive entrance.

Future work will focus on improving robustness to viewpoint changes and low-frame-rate videos, for example by adding more diverse training data and exploring tracking methods that make stronger use of temporal information.

\balance

\end{document}